\documentclass[10pt,twocolumn,letterpaper]{article}

\usepackage{cvpr}
\usepackage{times}
\usepackage{epsfig}
\usepackage{graphicx}
\usepackage{amsmath}
\usepackage{amssymb}

\usepackage[pagebackref=true,breaklinks=true,letterpaper=true,colorlinks,bookmarks=false]{hyperref}

 \cvprfinalcopy % *** Uncomment this line for the final submission

\def\cvprPaperID{****} % *** Enter the CVPR Paper ID here

\ifcvprfinal\fi
\begin{document}

%%%%%%%%% TITLE
\title{All-In-One: Facial Expression Transfer, Editing and Recognition Using A Single Network}

\author{Kamran Ali, Charles E. Hughes\\
Synthetic Reality Lab, Department of Computer Science\\
University of Central Florida, Orlando, Florida\\
{\tt\small kamran@knights.ucf.edu, ceh@cs.ucf.edu}
}

\maketitle
%\thispagestyle{empty}

%%%%%%%%% ABSTRACT
\begin{abstract}

   In this paper, we present a unified architecture known as Transfer-Editing and Recognition Generative Adversarial Network (TER-GAN) which can be used: 1. to transfer facial expressions from one identity to another identity, known as Facial Expression Transfer (FET), 2. to transform the expression of a given image to a target expression, while preserving the identity of the image, known as Facial Expression Editing (FEE), and 3. to recognize the facial expression of a face image, known as Facial Expression Recognition (FER). In TER-GAN, we combine the capabilities of generative models to generate synthetic images, while learning important information about the input images during the reconstruction process. More specifically, two encoders are used in TER-GAN to encode identity and expression information from two input images, and a synthetic expression image is generated by the decoder part of TER-GAN. To improve the feature disentanglement and extraction process, we also introduce a novel expression consistency loss and an identity consistency loss which exploit extra expression and identity information from generated images. Experimental results show that the proposed method can be used for efficient facial expression transfer, facial expression editing and facial expression recognition. In order to evaluate the proposed technique and to compare our results with state-of-the-art methods, we have used the Oulu-CASIA dataset for our experiments.
\end{abstract}

%%%%%%%%% BODY TEXT
\section{Introduction}

Facial Expression synthesis and manipulation is a challenging task because it requires the disentanglement of facial  expression  features  from  identity  information. It has recently gained a great deal of attention from the computer vision research community due to the exciting research challenges it offers apart from its many applications, e.g facial animation, human-computer interactions, entertainment and facial reenactment ~\cite{r1}. 

\begin{figure}[t]
    \centering
    \includegraphics[width=8.5cm,, height=7cm]{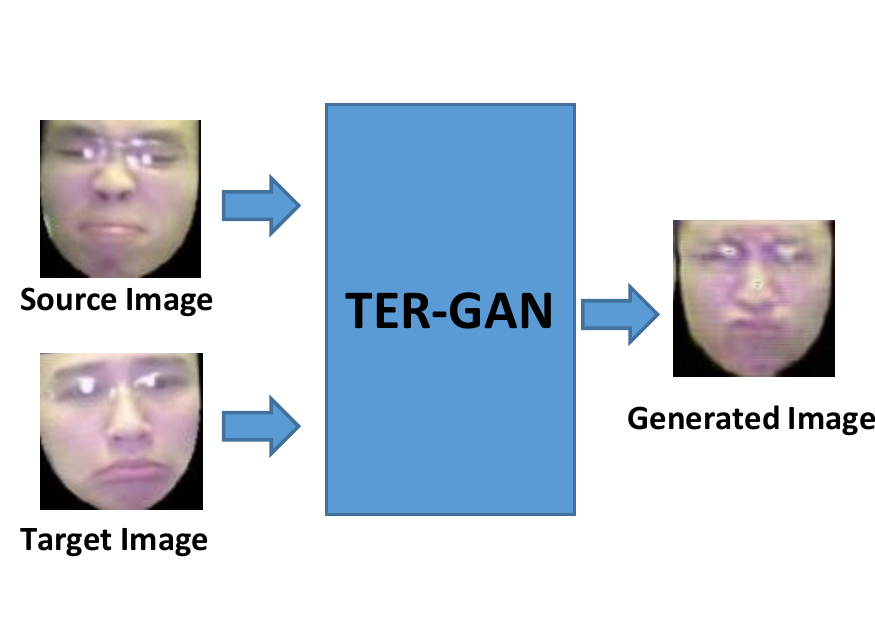}
    \caption{TER-GAN takes source image and target image as input and extracts expression information and identity information from each image respectively. These encoded information is then used to generate an expression image which contains the expression of source image while the identity of target image is preserved.}
    \label{fig:1}
\end{figure}

Many techniques have been developed for facial expression manipulation and editing. These techniques can be divided into two categories: graphic-based techniques ~\cite{r2}, ~\cite{r3}, ~\cite{r4} and generative methods ~\cite{r5}, ~\cite{r6}, ~\cite{r7},~\cite{r8}, ~\cite{r9}, ~\cite{r10}, ~\cite{r11}. In the first category, image warping is used to synthesize expression images by modeling the variations of a face during facial expressions. In the second category, deep generative models are used to generate synthesized expression images. In ~\cite{r8} an expression controller module is trained using a GAN-based architecture to generate expression images of various intensities. Similarly in ~\cite{r12}, a unified GAN framework is used to transfer expressions from one image domain to another. Kamran et al. ~\cite{r16} concatenated a one-hot identity vector with the expression information extracted from the encoder to generate an expression image. Pumarola et al.~\cite{r15} and Shao et al. ~\cite{r9} exploited Facial Action Units (AU), and the expression synthesis process is guided by the learned AU features. Similarly, in ~\cite{r13} and ~\cite{r14}, facial landmarks are used to produce synthesized expression images. 

Existing facial expression synthesis techniques have the capability to transform the expression of a given image; however, there are two main problems with these methods: 1. they require auxiliary information in the form of an expression code, facial landmarks and action unit information to synthesize an expression image and 2. many of these techniques fail to preserve the identity information of the given image, which is due to the fact that they fail to disentangle expression features from identity representation. Hence, during facial expression transfer process the identity information of the source image is usually leaked through the expression feature vector, which degrades the identity of generated images \cite{r46}.  Similarly, in \cite{r13} and \cite{r14}, it is very difficult to synthesize an expression image using the landmark information of source image with a different facial shape than the target face. To reduce the identity information leakage, in \cite{r16} and \cite{r8}, the expression synthesis process is conditioned on expression and identity codes, rather than using the extracted features. But it is a well known fact that the identity and expression representations are too rich to be represented by one-hot vectors.

In order to overcome the above problems, we propose a Transfer-Editing and Recognition Generative Adversarial Network (TER-GAN) to automatically and explicitly extract a disentangled expression representation from a source image and disentangled identity features from a target image in order to synthesize a photo-realistic expression image without requiring any auxiliary information such as expression or identity code, facial landmarks or action units, while preserving the identity of the target image. The overall framework of our proposed technique is shown in Figure \ref{fig:1}. TER-GAN has two main objectives: 1. to automatically extract disentangled expression features and identity features from a source image and a target image respectively, and 2. to synthesize a photo-realistic expression image containing the expression of the source image while preserving the identity of the target image. 

To achieve these objectives, we employ a Generative Adversarial Network (GAN) with an encoder-decoder based generator $G$. As opposed to previous generative model based expression synthesis and manipulation architectures ~\cite{r8}, ~\cite{r11}, ~\cite{r17}, we, instead of using just one encoder, employ two encoders $G_{es}$ and $G_{et}$ in our generator $G$. TER-GAN takes two images as input, source image $x_s$ and target image $x_t$. Encoder $G_{es}$ is aimed to encode expression representation $f(e)$ from source image $x_s$ and encoder $G_{et}$ is used to extract identity features $f(i)$ from target image $x_t$. The expression representation $f(e)$ is then concatenated with the identity feature $f(i)$: $f(x) = f(e) + f(i)$, and the concatenated feature vector $f(x)$ is then fed to the decoder $G_{de}$ to synthesize an expression image $\bar{x}$, which contains the expression of source image $x_s$ while preserving the identity of target image $x_t$. In order to further improve the quality of extracted expression and identity features, we make use of synthetic expression images along with real images, and introduce two adversarial losses at the output of each encoder: an adversarial expression consistency loss and an adversarial identity consistency loss. Our experimental results show that these two consistency losses help in extracting effective expression and identity features through which we can generate synthetic images that preserve the identity of the target image. Moreover, to generate more realistic synthesized expression images we use a multi-class classifier as our discriminator, $D$.
The main contributions of our paper are as follows:
\begin{itemize}
\item We present a novel unified architecture, Transfer-Editing and Recognition Generative Adversarial Network (TER-GAN) that can be used efficiently for three purposes: facial expression transfer, facial expression editing and facial expression recognition, without requiring any explicit expression or identity code or any other auxiliary information such as facial landmarks or action units to guide the synthesis process, while preserving the identity information.
\item Apart from the encoder-decoder architecture of TER-GAN, our adversarial expression and identity consistency losses also ensure that the expression and identity features are disentangled, and this disentanglement of features helps in synthesizing expression images that preserve the identity information of the target image.
\item In order to deal with small expression datasets, TER-GAN learns to extract expression and identity representations using the information contained in synthesized images as well.
\item We show that the disentangled expression embedding learned by TER-GAN can be effectively used for facial expression recognition.

\end{itemize}
%-------------------------------------------------------------------------

\begin{figure*}[ht!]
\centering
%[scale=1, width=.01\textwidth]
\includegraphics[width=16cm,, height=7.5cm]{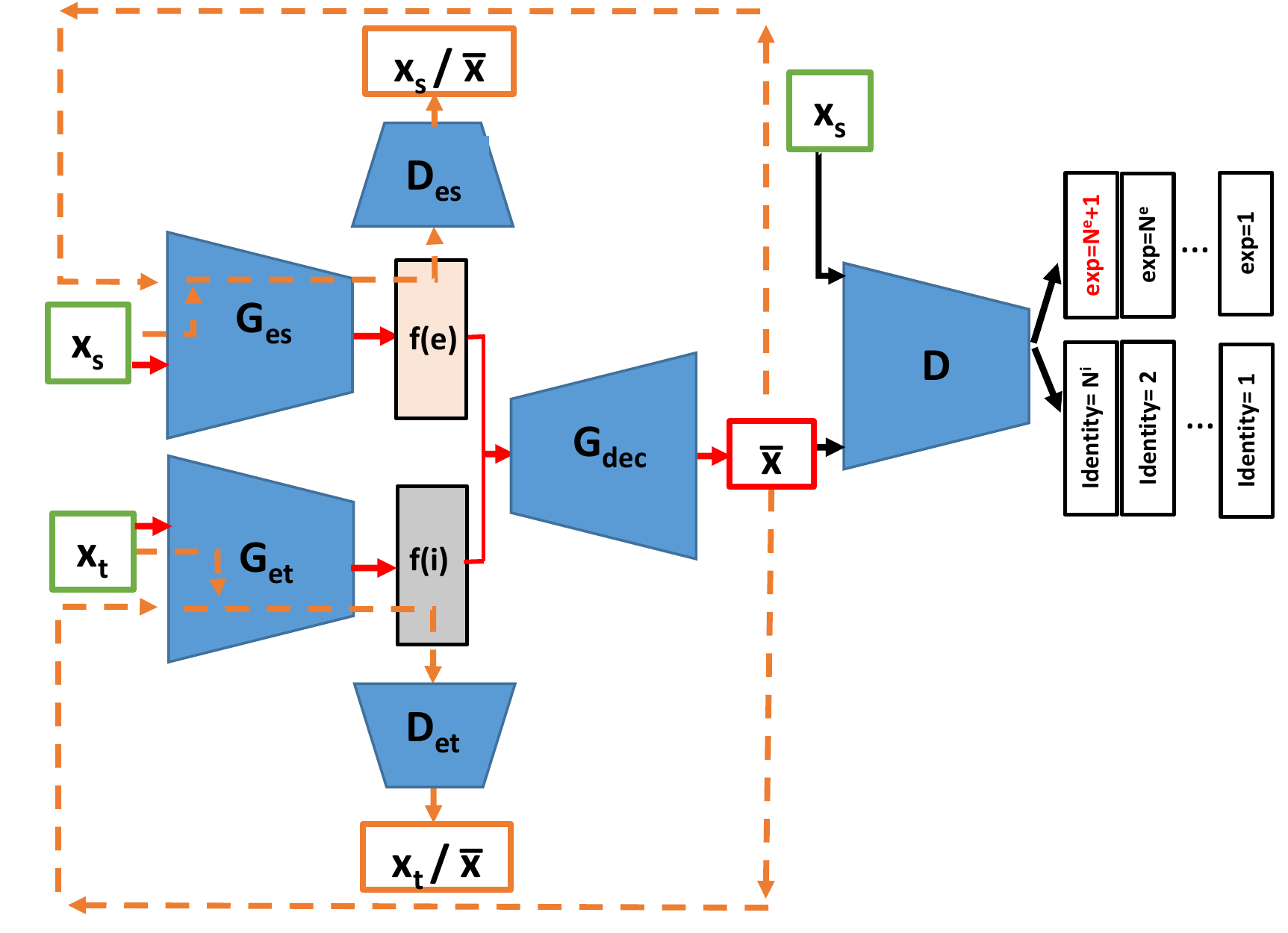}
\caption{The over-all architecture of our TER-GAN (best viewed in color)}
\label{fig:2}
\end{figure*}

%---------------------------------------------------------------
\section{Related Work}

In this section we first review previous facial expression synthesis and manipulation techniques, followed by discussing conventional feature disentanglement techniques.
%-------------------------------------------------------------------------
\subsection{Facial Expression Manipulation}
There are many facial expression manipulation techniques proposed in the literature, some of which combine computer graphics methods such as 2D or 3D image warping ~\cite{r33}, flow mapping ~\cite{r2} and image rendering ~\cite{r34} with computer vision algorithms to generate synthesized expression images. Although these techniques are able to produce photo-realistic high resolution synthesized images, the main problem with these approaches is that they suffer from high computation cost. In contrast to conventional facial expression synthesis techniques, recently proposed methods are mostly based on conditional Generative Adversarial Networks (cGANs).

Some earlier techniques such as \cite{r35}, \cite{r36}, and \cite{r37} used deterministic target expressions as one-hot vectors and generated synthesized images conditioned on discrete facial expressions. While the generated images are of better quality, these techniques are only able to generate discrete expressions. To overcome this problem, Ding et al. \cite{r8} proposed an Expressive GAN (ExprGAN) to synthesize an expression image conditioned on a real-valued vector that contains more complex information such as intensity variation. Similarly in \cite{r13} and \cite{r38}, the image synthesis process is conditioned on geometry information in the form of facial landmarks. Choi et al. \cite{r39} proposed the StarGAN method to employ domain information and generate an image into a corresponding domain. In \cite{r39}, Royer et al. presents XGAN to translate attributes from one domain to another by using an adversarial classifier on top of encoded layers, but their architecture is too simple to generate photo-realistic expression images. In another work, Pumarola et al. \cite{r37} used AUs as a conditional label to synthesize an expression image. All of these proposed techniques require explicit expression, AU and landmark information to guide the expression synthesis process.
\subsection{Feature Disentanglement}
Many techniques have been developed in the past to disentangle an image into its different representations. These techniques are based on the idea of learning by reconstruction, and therefore, often involve encoder-decoder structure coupled with GANs. For instance Tran et al. \cite{r40} proposed a disentangled representation learning GAN for pose-invariant face recognition in which the face identity representation is disentangled from the pose information by explicitly providing the pose information to the generator. Similarly, Lee et al. \cite{r41} generated photo-realistic images from various domains by disentangling the factors of variations in an input image. Shu et al. \cite{r42} proposed an unsupervised generative model to disentangle shape from appearance information. All of these methods learn the disentangled representation by explicitly providing information about other (uninterested) factors of variation that constitute an image. In contrast, our method automatically extracts the factors of variations from input images using two augmenting feature learning techniques, i.e learning by reconstruction employing an encoder-decoder based GAN set-up and by applying adversarial expression consistency loss and adversarial identity consistency loss at each encoder. 
\section{Overview of TER-GAN}
The main objective of TER-GAN is twofold: to extract descriptive, discriminative and disentangled expression and identity representations from input images, and secondly to generate synthesized expression images using the extracted expression and identity information in such away that the expression and identity information of input images are preserved. To do this, as opposed to previous facial expression synthesis architectures, where expression or identity information is explicitly provided in the form of expression or identity codes, our TER-GAN uses two encoders to automatically extract expression and identity information. It has been reported in the literature that face images lie on a manifold ~\cite{r17}, therefore, we argue that representing a face image with an identity code in the form of a one-hot vector is not enough to capture fine details of a target face. Similarly, in order to represent various intensities of a facial expression, using just an expression code is not sufficient to generate wide range of expression intensities. In order to learn efficient expression and identity representation, we, in addition to real images, use the synthetic expression images generated by the decoder of our generator. Thus, in this manner joint expression-invariant identity embedding and identity-invariant expression embedding are learned using two additional adversarial losses on top of representation layers at each encoder, apart from  the adversarial loss imposed on the overall generator.
\subsection{Network Architecture}
The input to TER-GAN is a source image $x_s$ and a target image $x_t$. These two images are fed to two different encoders $G_{es}$ and $G_{et}$, where $G_{es}$ aims to map source image $x_s$ to an expression representation $f(e)$, while $G_{et}$ is used to project the target image $x_t$ to an identity embedding $f(i)$. The concatenation of the two embeddings:  $f(x) = f(e) + f(i)$, bridges the two encoders with a decoder $G_{de}$. The objective of decoder $G_{de}$ is to synthesize an expression image $\bar{x}$ having the expression $e$ of the source image and the identity $i$ of the target image: $\bar{x} = G_{de} (f (x))$. To further improve the quality of synthesized images, an adversarial loss is imposed on generator $G$ by using a multi-class CNN as our discriminator $D$. In order to generate synthetic images with desired expressions and identities, our discriminator $D$ performs identity and expression classification, apart from classifying between real and fake images. The overall architecture of the proposed TER-GAN is shown in Figure \ref{fig:2}. Two additional discriminators $D_{es}$ and $D_{et}$ are used with $G_{es}$ and $G_{et}$, respectively, to learn an identity-invariant expression embedding $f(e)$ at the FC layer of encoder $G_{es}$ and an expression-invariant identity embedding $f(i)$ at the FC layer of encoder $G_{et}$. The adversarial learning scheme of identity-invariant expression embedding $f(e)$ and expression-invariant identity embedding $f(i)$ is shown with a brown-colored dashed line in in Figure \ref{fig:2}.
\subsubsection{Discriminator}
The discriminator $D$ of TER-GAN is a multi-task CNN that aims for three objectives: 1. to classify between real and fake images, 2. to classify facial expressions, and 3. to classify the identities of expression images. To achieve these objectives, discriminator $D$ is divided into two parts: $D = [D^e, D^i]$, where $D^e \in R^{N^{e+1}}$ corresponds to the part of $D$ that is used for the classification of expressions i.e $N^e$ denotes the number of expressions, in our case it represents six basic expressions, and an additional dimension is used to differentiate between real and fake images. Similarly,  $D^i \in R^{N^i}$ is the part of $D$ that is used to classify the identities of expression images, where $N^i$ denotes the number of identities. The overall objective function of our discriminator $D$ is given by the following equation:
\begin{align}
\underset{D}{\mathrm{max}}\mathcal{L_D}(D,G)={}&\underset{x_t,y_t\sim p_t(x_t,y_t)}{E_{{x_s,y_s\sim p_s(x_s,y_s)},}}[\log({D_{y_s^e}^e}(x_s)+\notag\\
&\log({D_{y_t^{i}}^{i}}(x_t)]+\notag\\
&\underset{x_t,y_t\sim p_t(x_t,y_t)}{E_{{x_s,y_s\sim p_s(x_s,y_s)},}}[\log({D_{N^e+1}^e}{(G(x_s,x_t))}]\notag\\
\label{eq:1}
\end{align}
The first part of equation \ref{eq:1} represents the objective of $D$ to maximize the probability of source image $x_s$ and target image $x_t$ to be classified to its true expression label $y_s$ and true identity label $y_t$, respectively. While the second part of the function corresponds to the objective of $D$ to maximize the probability of classifying $\bar{x}$ as a fake image.
\subsubsection{Generator}
In previous facial expression synthesis and manipulation methods ~\cite{r8},~\cite{r18},~\cite{r19} one encoder is used to extract feature information from an input image, while a conditional code is explicitly fed to the network to guide the facial expression synthesis process. However, in TER-GAN, the main objective of $G$ is to efficiently extract expression and identity representation from source image $x_s$ and target image $x_t$, respectively, and to generate an image $\bar{x}$ to fool $D$ to classify it to the expression of $x_s$ and the identity of $x_t$. Therefore, the generator $G$ in TER-GAN consists of two encoders and a decoder: $G = (G_{es},G_{et},G_{de})$.  The objective function of $G$ is given by the following equation:

\begin{align}
\underset{G}{\mathrm{max}}\mathcal{L_G}(D,G)={}&\underset{x_t,y_t\sim p_t(x_t,y_t)}{E_{{x_s,y_s\sim p_s(x_s,y_s)},}}[\log({D_{y_s^e}^e}(G(x_s,x_t))+\notag\\
&\log({D_{y_t^{i}}^{i}}(G(x_s,x_t))]
\end{align}

The adversarial loss is given as below:

\begin{align}
\underset{G,D}{\mathrm{max}}\mathcal{L}_{adv}(D,G)={}&\mathcal{L}_G + \mathcal{L}_D 
\end{align}

In order to improve the capability of both of our encoders to extract identity-invariant expression features and expression-invariant identity features, we introduce two additional adversarial losses on top of the representation layer of each encoder: adversarial expression consistency loss at encoder $G_{es}$ and adversarial identity consistency loss at $G_{et}$. 

\textit{Encoder $G_{es}$}: The main objective of encoder $G_{es}$ is to extract expression representation $f(e)$ from input source image $x_s$. To achieve this goal, apart from employing learning by reconstruction phenomena, we propose another adversarial expression consistency loss at encoder $G_{es}$, which does not require any paired data and helps in learning an identity-invariant expression representation in a self-supervised manner. Specifically, since the input source image $x_s$ and the generated image $\bar{x}$ share the same expression information but have different identities, we leverage these two images to learn an identity-invariant expression embedding. To do this, a discriminator $D_{es}$ is trained on top of expression embedding $f(e)$, to classify the encoded features to be extracted from $x_s$ or $\bar{x}$. To learn an identity-invariant expression embedding $f(e)$, discriminator $D_{es}$ strives to maximize its classification accuracy, while encoder $G_{es}$ is trained to confuse discriminator $D_{es}$ by minimizing its accuracy. The optimization function is given by the equation below:
\begin{align}
\underset{G_{es}}{\mathrm{min}}\quad\underset{D_{es}}{\mathrm{max}}\mathcal{L}_{D_{es}}={}&{E_{{x_s\sim p_s(x_s)},}}\mathcal{L}(1, D_{es}(G_{es}(x_s)))+\notag\\
&{E_{{\bar{x}\sim p_{\bar{x}}(\bar{x})}}}\mathcal{L}(2, D_{es}(G_{es}(\bar{x})))
\end{align}
Where $\mathcal{L}$ denotes a cross-entropy loss.

\textit{Encoder $G_{et}$}: The target image $x_t$ is fed to encoder $G_{et}$, which extracts identity representation $f(i)$ for image synthesis. The target image $x_t$ can have any expression or it can be a neutral image, since it is only used for getting identity information. Therefore, to extract an expression-invariant identity representation, we employ an adversarial identity consistency loss on top of identity representation layer $f(i)$ at encoder $G_{et}$. The synthesized image $\bar{x}$, which has the same identity as $x_t$ is fed to encoder $G_{et}$ along with the input target image $x_t$ to learn the expression-invariant identity embedding. This goal is achieved by using a discriminator $D_{et}$ on top of identity embedding $f(i)$, which is trained to recognize the encoded identity representation $f(i)$ as coming from $x_t$ or $\bar{x}$. The discriminator $D_{et}$ strives to maximize its classification accuracy while the encoder $G_{et}$ is aimed to confuse discriminator $D_{et}$ by minimizing its accuracy. The optimization function is given by the equation below:
\begin{align}
\underset{G_{et}}{\mathrm{min}}\quad\underset{D_{et}}{\mathrm{max}}\mathcal{L}_{D_{et}}={}&{E_{{x_t\sim p_t(x_t)},}}\mathcal{L}(1, D_{et}(G_{et}(x_t)))+\notag\\
&{E_{{\bar{x}\sim p_{\bar{x}}(\bar{x})}}}\mathcal{L}(2, D_{et}(G_{et}(\bar{x})))
\end{align}
Where $\mathcal{L}$ denotes a cross-entropy loss.

\textit{Decoder $G_{de}$}: The input to decoder $G_{de}$ is a concatenation of  $f(e)$ and $f(i)$: $f(x) = f(e) + f(i)$, through which $G_{de}$ will generate a synthesized image having the expression encoded in $f(e)$ and the identity information represented by $f(i)$. For this purpose, two pixel-wise reconstruction losses are used: 1. an identity reconstruction loss between input target image $x_t$ and output image $\bar{x}$ and 2. an expression reconstruction loss between input source image $x_s$ and output image $\bar{x}$. 
\begin{align}
\underset{G_{es},G_{et},G_{de}}{\mathrm{min}}\mathcal{L}_{irec}={}&L_1(G_{de}(G_{es}(x_s),G_{et}(x_t)),x_t)
\end{align}
\begin{align}
\underset{G_{es},G_{es},G_{de}}{\mathrm{min}}\mathcal{L}_{erec}={}&L_1(G_{de}(G_{es}(x_s),G_{et}(x_t)),x_s)
\end{align}
However, it is known that pixel-level metrics are not very optimal for the purpose of image comparison, especially when dealing with semantics level comparison ~\cite{r20}, ~\cite{r8}. Therefore, to further preserve the expression information between source image $x_s$ and synthesized image $\bar{x}$, a pre-trained version of encoder $G_{es}$ is used to enforce expression similarity in feature space: 
\begin{align}
\underset{G_{es},G_{et},G_{de}}{\mathrm{min}}\mathcal{L}_{ef}={}&\sum_l\omega_{1l}L_1(h_{1l}(G_{de}(G_{es}(x_s),G_{et}(x_t)),h_{1l}(x_s))
\end{align}
where $h_{1l}$ represents the $l_{th}$ layer feature maps extracted from the pre-trained version of $G_{es}$ and $\omega_{1l}$ denotes the $l_{th}$ layer's weight. The activations at all five convolutional layers of the network are used.

Similarly, to further preserve the identity information between target image $x_t$ and synthesized image $\bar{x}$, the pre-trained version of encoder $G_{et}$ is used to extract identity features from various inter-mediate layers. The feature-level identity preserving loss is given by the following equation:
\begin{align}
\underset{G_{es},G_{et},G_{de}}{\mathrm{min}}\mathcal{L}_{if}={}&\sum_l\omega_{2l}L_1(h_{2l}(G_{de}(G_{es}(x_s),G_{et}(x_t)),h_{2l}(x_t))
\end{align}
\subsubsection{Total Loss}
The overall objective function of TER-GAN is the weighted sum of all the losses discussed in the previous sections:
\begin{align}
\underset{G_{es},G_{et}}{\mathrm{min}}\quad\underset{D_{et}}{\mathrm{max}}\mathcal{L}_{TER-GAN}={}&\lambda_1\mathcal{L}_{if} + \lambda_2\mathcal{L}_{ef} + \lambda_3\mathcal{L}_{irec}+\notag\\
&\lambda_4\mathcal{L}_{erec}+\lambda_5\mathcal{L}_{D_{et}}+\lambda_6\mathcal{L}_{D_{es}}+\notag\\
&\lambda_7\mathcal{L}_{adv}
\label{eq:2}
\end{align}
\subsubsection{Training the Network}
Since the efficiency of learning an identity-invariant expression embedding and an expression-invariant identity embedding depends on the quality of the generated images as well, TER-GAN, having multiple loss functions, is trained employing a curriculum strategy ~\cite{r8}, ~\cite{r46} with two training stages. In the first stage of training, the encoder $G_{es}$ is pre-trained in a supervised manner to classify facial expressions. Similarly, the encoder $G_{et}$ is pre-trained in a fully supervised way to recognize identities. The classification layers of these two networks are then discarded and the rest of the remaining networks are attached to the over-all architecture of TER-GAN. In the second stage of training, the entire TER-GAN architecture is trained in two steps of updating the network parameters. In the first step, the expression $f(e)$ and identity $f(i)$ features are extracted from the source image $x_s$ and the target image $x_t$ by the two encoders $G_{es}$ and $G_{et}$, and these two features are concatenated into $f(x)$ and fed to decoder $G_{de}$ to synthesize an expression image $\bar{x}$. In the second step, the generated output image $\bar{x}$ is fed to two encoders along with their corresponding input images to further improve the quality of $f(e)$ and $f(i)$, and to learn in an adversarial manner an identity-invariant expression representation $f(e)$ and expression-invariant identity embedding $f(i)$. 
\begin{figure}[t]
    \centering
    \includegraphics[width=7.5cm,, height=4.5cm]{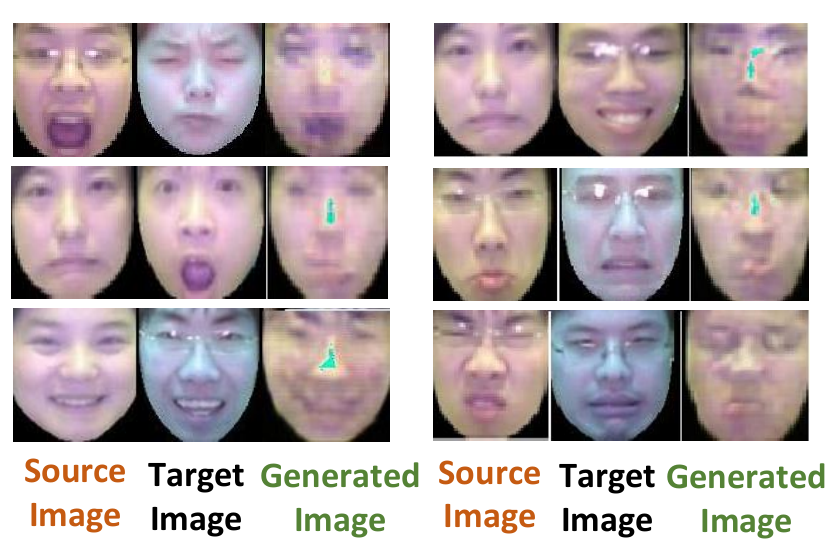}
    \caption{TER-GAN encodes expression information from source image, and identity information from target image, and generates an output expression image having the expression of source image while preserving the identity of the target image.}
    \label{fig:3}
\end{figure}
\begin{figure}[t]
    \centering
    \includegraphics[width=8.6cm,, height=11.5cm]{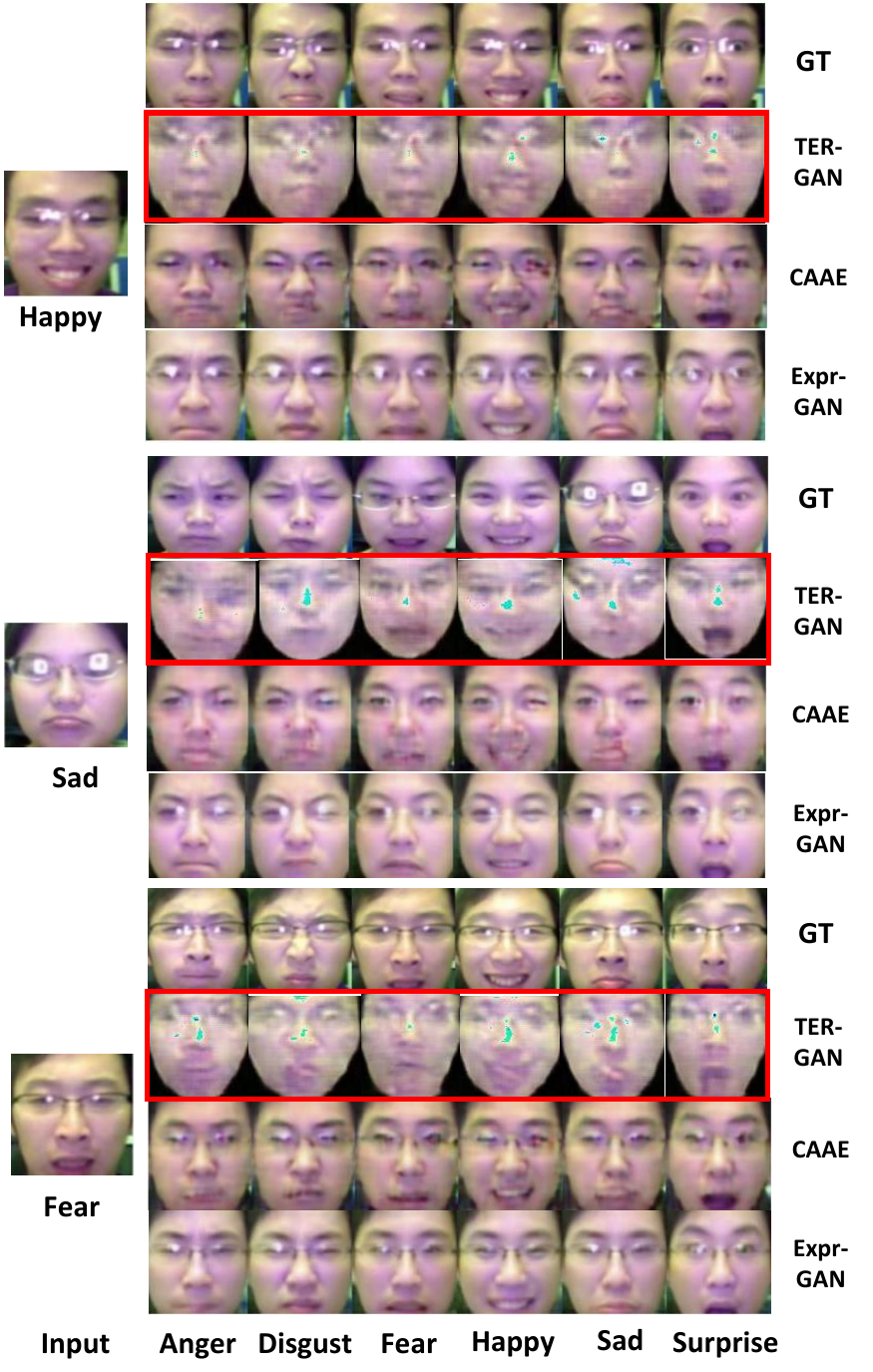}
    \caption{Comparison (visual) of facial expression editing techniques. For each input (target image), the corresponding synthetic expression images are generated. We compare our results with CAAE and Expr-GAN.  }
    \label{fig:4}
\end{figure}

The objective of training TER-GAN is to minimize equation \ref{eq:2}. Specifically, the adversarial losses $\mathcal{L}_{D_{es}}$, $\mathcal{L}_{D_{et}}$ and $\mathcal{L}_{adv}$ lead to a min-max optimization problem, resulting in our employing a gradient reversal layer ~\cite{r47}, ~\cite{r20} into TER-GAN architecture. The gradient reversal layer is implemented between $D_{es}$ and expression embedding $f(e)$ in order to perform adversarial training scheme for $\mathcal{L}_{D_{es}}$. The gradient reversal layer does not affect the forward pass during training, but it is used to invert the gradient sign during back-propagation to practically implement the min-max training scheme. The gradient reversal layer is also used between $D_{et}$ and identity embedding $f(i)$ in order to perform adversarial training scheme for $\mathcal{L}_{D_{et}}$.
\section{Experiments}
In this section we first describe the implementation details, followed by presenting the experimental results of TER-GAN and then we demonstrate the applications of TER-GAN and its the ability to perform facial expression  transfer and facial expression editing and to support facial expression recognition.
\subsection{Implementation Details}
The proposed technique is evaluated on the widely used Oulu-CASIA ~\cite{r22} dataset. Initially Convolutional Experts Constrained Local Model (CE-CLM) is used to detect facial landmarks to perform face detection and face alignment. To avoid overfitting due to using small dataset, data augmentation is performed to increase the number of images in the training dataset. From each image, five $75\times 75$ samples are extracted from the center and four corner locations. Image rotation is then applied on each of those $75\times 75$ cropped samples using four angles: $-6^\circ$, $-3^\circ$, $3^\circ$, $6^\circ$. Horizontal flipping is then applied on each rotated image, and thus, the size of the dataset is increased 5 times the original dataset size after data augmentation.

\begin{table}
\begin{center}
\begin{tabular}{|l| c| c|} 
 %\hline
 %\multicolumn{4}{|c|}{Country List} \\
 \hline
 Method&Setting&Accuracy\\
 \hline
 LBP-TOP\cite{r32}   & Dynamic    &68.13\\
 HOG 3D\cite{r31}&   Dynamic  & 70.63\\
 STM-Explet\cite{r30}    &Dynamic & 74.59\\
 Atlases\cite{r29}&   Dynamic  & 75.52\\
 DTAGN\cite{r26}&   Dynamic  & 81.46\\
 FN2EN\cite{r27}&   Static  & 87.71\\
 PPDN\cite{r28}&   Static  & 84.59\\
 DeRL\cite{r25}&   Static  & 88.0\\
 \hline
 CNN(baseline)& Static  & 73.14\\
 \textbf{TER-GAN(Ours)}& Static  & \textbf{89.65}\\
 \hline
\end{tabular}
\end{center}
\caption{Oulu-CASIA: Accuracy for six expressions classification.}
\label{table:1}
\end{table}

TER-GAN is initially pre-trained on the BU-4DFE ~\cite{r23} dataset, which consists of 60,600 images from 101 identities. Six image sequences are captured for each identity, and these six sequences correspond to six basic expressions. Each of these sequences are arranged in such a way that it starts from a neutral expression, reaches the peak expression in the middle, and then again ends at a neutral expression. The middle peak expression images are extracted to construct the dataset.
\begin{figure*}[ht!]
\centering
%[scale=1, width=.01\textwidth]
\includegraphics[width=16.5cm,, height=6.3cm]{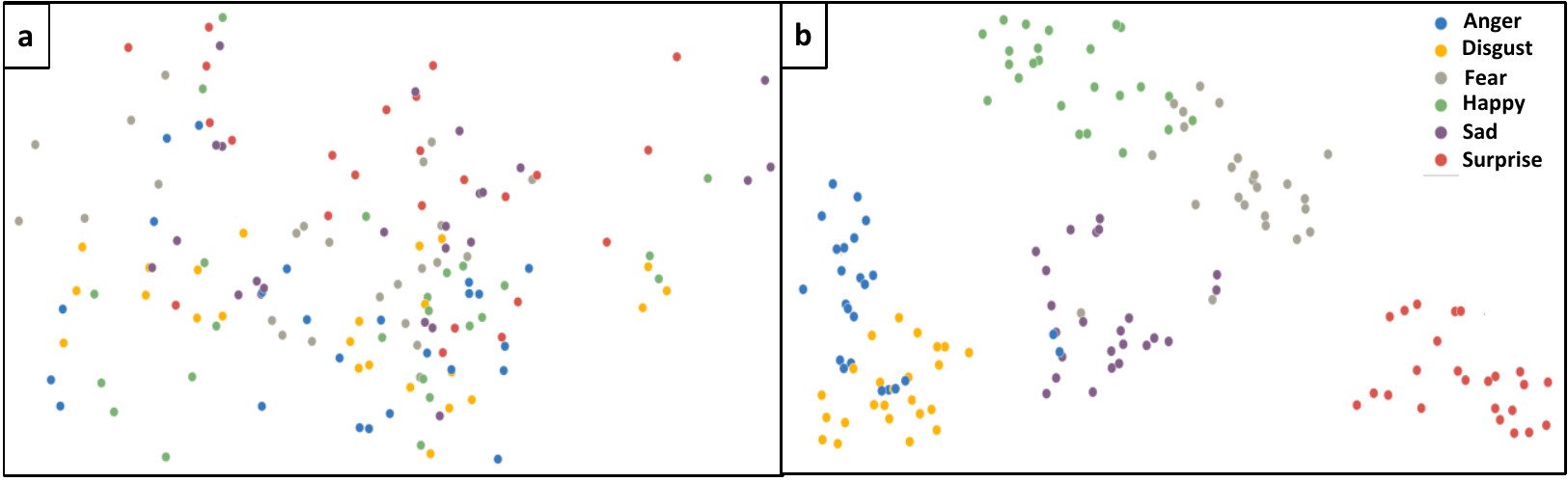}
\caption{Expression feature space. Each color represents a different expression. Fig. 5a shows the expression distribution of features obtained from pre-trained encoder $G_{es}$ (CNN baseline). Fig 5b. depicts the expression distribution of features obtained from encoder $G_{es}$ after training in a TER-GAN set-up. (Best viewed in color)}
\label{fig:5}
\end{figure*}

The pre-trained TER-GAN is then fine-tuned on the Oulu-CASIA (OC) dataset. OC dataset contains 480 video sequences captured under three different illumination conditions using two different cameras. In this experiment, only images captured under strong condition with VIS camera are used. There are 80 identities in the OC dataset, and each identity has six video sequences, corresponding to six basic expressions. Each video sequence starts with a neutral image and ends at the peak expression image. In this experiment the last three frames of each sequence are selected to construct the dataset. An identity-independent training-testing split is formed to evaluate the proposed method.

The architecture of both encoders, $G_{es}$ and $G_{et}$, is designed based on five downsampling blocks consisting of a $3\times3$ stride 1 convolution. The number of channels are 64, 128, 256, 512, 1024 and one 30-dimensional FC layer for expression feature vector $f(e)$, and a 50-dimensional identity representation $f(i)$, constitute $G_{es}$ and $G_{et}$, respectively. The decoder $G_{de}$ is built on five upsampling blocks containing a $3\times3$ stride 1 convolution. The number of channels are 512, 256, 128, 64 and 3. As opposed to previous GAN architectures with a multi-task CNN based discriminator \cite{r18}, \cite{r16}, in TER-GAN, the discriminator $D$ is designed in such a way that the initial downsampling convolutional layers and a FC layer are shared between $D^e$ and $D^i$ in order to reduce the computation cost. More specifically, four CNN blocks with 16, 32, 64, 128 channels and a 1024-dimensional FC layer are shared between the two parts. It is then divided into two branches, where, each branch has two additional FC layers with 512 and 256 channels. $D^e$ then has an expression classification layer and $D^i$ has an identity classification layer. The architectures of $D_{es}$ and $D_{et}$ are the same, which consists of three FC layers with channels 32, 16 and 1. 

TER-GAN is trained using the Adam optimizer \cite{r24}, with a batch size of 64 and learning rate of 0.0002. The values of parameters are empirically set to $\omega_{11} = \omega_{21} = 0.5$, $\omega_{12} = \omega_{22} = 0.6$, $\omega_{13} = \omega_{23} = 0.7$, $\omega_{14} = \omega_{24} = 0.88$, and $\omega_{15} = \omega_{25} = 0.99$. Similarly, the weights of the total loss are set empirically as $\lambda_1 = 1 $, $\lambda_2 = 1 $, $\lambda_3 = 1$, $\lambda_4 = 1$,  $\lambda_5 = 0.3$,  $\lambda_6 = 0.3$, and  $\lambda_7 = 0.5$.
\subsection{Facial Expression transfer}
In this section, we demonstrate our model's ability to transfer facial expressions from source image $x_s$ to target image $x_t$. As opposed to previous methods \cite{r8}, where, a separate expression classifier is used to extract expression label in order to transfer facial expression from one image to another image, in TER-GAN, the facial expression transfer task is performed in an end-to-end manner. To do this, $x_s$ is fed to encoder $G_{es}$ and $x_t$ is input to encoder $G_{et}$ to extract expression information from $x_s$ and identity features from $x_t$ respectively. The expression information is then concatenated with the identity features, and the concatenated feature vector is fed to decoder $G_{de}$ to synthesize an expression image having the expression of $x_s$ and containing the identity of $x_t$. Figure \ref{fig:3} shows that TER-GAN transfers the facial expressions from source images quite accurately, while also preserving the target identities specified by target images.  
\subsection{Facial Expression Editing}
In this section we demonstrate the capability of our proposed TER-GAN to edit the expression of a given image. Different from previous facial expression editing methods, like in \cite{r8}, where the expression code is explicitly fed to the network, in TER-GAN, the expression information is extracted from another expression image (source image) to encode more valuable expression information than just an expression label. In this experiment, the identity feature $f(i)$ is extracted from the given image (target image, $x_t$) by inputting it to encoder $G_{et}$, while the expression information $f(e)$ is automatically extracted from another image of the same identity but with a different expression (source image $x_s$). $f(e)$ and $f(i)$ are then concatenated and the concatenated feature vector is then fed to the decoder $G_{de}$ to generate a synthetic image $\bar{x}$, which has the expression information taken from $x_s$, while preserving the identity information. The experimental results are shown in Figure \ref{fig:4}, where the first column corresponds to the input image (target image), while the ground truth images in top row in the right column represent the source images, $x_s$, which are used to extract the corresponding expression information. Although, our main objective is different than the previous methods \cite{r8}, \cite{r43}, and the network architecture of TER-GAN has more functionalities, and is more complex than the models proposed in \cite{r8} and \cite{r43}, we, for the sake of comparison, compare our results with \cite{r8} and \cite{r43}. As it can be seen in Figure \ref{fig:4}, our TER-GAN can not only synthesize an image of the desired expression, but the identity information is preserved more in our case than in \cite{r8} and \cite{r43}. To reduce the computational cost and model complexity, we have not used any variation regularization \cite{r44} method as used in \cite{r8} to reduce spike artifacts on the reconstructed images, which, we believe, if incorporated in TER-GAN, will enhance the visual quality of our synthesized images as well.

\subsection{Facial Expression Recognition}
In this section we demonstrate the ability of TER-GAN to efficiently disentangle the expression information of any expression image from its identity information. One of the major issues with conventional FER techniques is that the representation used for facial expression recognition contains identity information as well as the expression information and, as a result, the performance of FER degrades on unseen identities during real-time applications. Therefore, in order to obtain identity free expression information, the encoder $G_{es}$ of TER-GAN is detached from the rest of the architecture after training and is used to perform facial expression recognition. Specifically, an expression image is fed to the detached encoder $G_{es}$ and the output expression representation $f(e)$ is extracted. This feature vector is then fed to a shallow classifier for facial expression recognition. 

To evaluate the performance of the proposed disentangled facial expression recognition technique, we conducted an eight fold cross validation on the Oulu-CASIA dataset. Table  \ref{table:1} shows the average accuracy obtained using the proposed technique. The reported results show that TER-GAN outperforms state-of-the-art techniques including GAN-based methods like DeRL ~\cite{r25} and deep CNN-based techniques like DTAGN-Joint ~\cite{r26}, FN2EN ~\cite{r27}, and PPDN ~\cite{r28}. Although, we are using only images to extract expression information for FER, our method out-performs techniques like DTAGN\cite{r26}, Atlases\cite{r29}, STM-Explet\cite{r30}, HOG 3D\cite{r31} and LBP-TOP\cite{r32} that exploit temporal information of video sequences.
\subsection{Expression Feature Visualization}
In this part, we demonstrate that the expression representation $f(e)$ learned by our proposed TER-GAN is disentangled from identity information. To do this, we first extract the expression feature vector $f(e)$ from encoder $G_{es}$, and then employ t-SNE ~\cite{r45} to project the 30-dim feature vector $f(e)$ on a two dimensional space for visualization purpose. The 2d expression feature space is shown in Figure \ref{fig:5}. For the sake of comparison with our CNN baseline, we conduct two experiments to show that the expression representation learned by encoder $G_{es}$, when trained in a TER-GAN set-up, is disentangled from identity information. In the first experiment, we extract 30-dim expression features from our CNN baseline network (pre-trained encoder $G_{es}$), and visualize this in a 2d feature space using t-SNE. The result of our first experiment is shown in Figure \ref{fig:5}(a). It can be seen that the expression features are all entangled with each other in the expression feature space, which clearly indicates that the CNN baseline model fails to disentangle expression information from identity features. In the second experiment, we employ encoder $G_{es}$ trained in an end-to-end manner in a TER-GAN set-up, to extract expression representation $f(e)$, and project it to 2d feature space using t-SNE. The result of the second experiment is shown in Figure \ref{fig:5}(b). We can see that the expression features are organised in the form of six clusters, corresponding to six basic expressions, which indicates that the expression information is effectively disentangled from identity information. 
\section{Conclusion}
In this paper we have proposed a unified expression transfer, editing and recognition architecture, TER-GAN, which has two objectives: 1). to extract efficient and disentangled expression and identity features from input images, and 2). to employ the extracted expression and identity representations for realistic looking expression synthesis that preserves the identity information of the target (given) image. This goal is achieved by explicitly encoding the expression information from a source image and extracting identity information from a target image by using two different dedicated encoders, and these two feature vectors are than combined to generate an expression image by employing the decoder part of TER-GAN. In order to further improve the expression and identity feature extraction process, we have introduced novel expression and identity consistency losses. Experimental results show that the proposed method can be used for efficient facial expression transfer and facial expression editing, and the disentangled feature representation can be used for facial expression recognition.
%-------------------------------------------------------------------------

%-------------------------------------------------------------------------
{\small
\bibliographystyle{ieee_fullname}
\bibliography{egbib}
}

\end{document}